\documentclass[default,iicol]{sn-jnl-revised}

\jyear{2023}%

\theoremstyle{thmstyleone}%
\theoremstyle{thmstyletwo}%

\theoremstyle{thmstylethree}%

\usepackage{bbold}
\usepackage{multirow}
\usepackage{bm}
\usepackage{arydshln}
\usepackage{xspace}
\usepackage{url}

\makeatletter
\newcommand{\figcaption}[1]{\def\@captype{figure}\caption{#1}}
\newcommand{\tblcaption}[1]{\def\@captype{table}\caption{#1}}
\newcommand\customparagraph[1]{\vspace{0.7em}\noindent\textbf{#1}}
\makeatother

\DeclareRobustCommand\red{\textcolor{red}}

\DeclareRobustCommand\green{\color[rgb]{0,0.5,0}}

\def\tbf{\textbf}
\def\tit{\textit}

\makeatletter
\DeclareRobustCommand\onedot{\futurelet\@let@token\@onedot}
\def\@onedot{\ifx\@let@token.\else.\null\fi\xspace}
\def\eg{\emph{e.g}\onedot} 
\def\ie{\emph{i.e}\onedot} 
 
 \def\vs{\emph{vs}\onedot}
 
\def\etal{\emph{et al}\onedot}

\makeatother

\def\tblannot{
\begin{table*}[tp]
\caption{Taxonomy of methods for annotating 3D hand poses. We categorize the annotation methods as manual, synthetic-model-based, hand-marker-based, and computational annotation.}
\footnotesize{
\scalebox{0.82}[1]{
\begin{tabular}{llcccrrrccl}
Annotation                   & Dataset                                               & Year & Modality & Resolution       & \#Frame & \begin{tabular}[c]{@{}l@{}}\#Subj./\\\#Obj.\end{tabular} & \#View & Motion                                                        &\begin{tabular}[c]{@{}l@{}}Obj.\\pose\end{tabular}                                                & Sensor/Engine                          \\\hline\rule[0mm]{0mm}{3mm}
\tbf{Manual}                 & MSRA~\cite{qian:cvpr14}                               & 2014 & Depth    & 320$\times$240   & 2K     & 6 / -       & 1 (3rd)  & \red{$\times$}                                          & \red{$\times$}                                         & Creative Senz3D        \\ 
                             & Dexter+Object~\cite{sridhar:eccv16}                   & 2016 & RGB-D    & 640$\times$480   & 3K     & 2 / 2       & 1 (3rd)  & {\green{\checkmark}}                                          & {\green{\checkmark}}                                          & Creative Senz3D                   \\ 
                             & EgoDexter~\cite{mueller:iccv17}                       & 2017 & RGB-D    & 640$\times$480   & 3K     & 4 / -       & 1 (1st)  & {\green{\checkmark}}                                          & \red{$\times$}                                                & RealSense SR300             \\
\tbf{Synthetic}              & SynthHands~\cite{mueller:iccv17}                      & 2017 & RGB-D    & 640$\times$480   & 220K   & - / 7       & 5 (1st)  & \red{$\times$}                                                & \red{$\times$}                                                & Unity                                \\
                             & RHD~\cite{zimmermann:iccv17}                          & 2017 & RGB      & 320$\times$320   & 43K    & 20 / -       & 1 (3rd)  & \red{$\times$}                                                & \red{$\times$}                                                & Unity                             \\
                             & GANerated~\cite{mueller:cvpr18}                       & 2018 & RGB      & 256$\times$256   & 331K   & - / -       & 1 (3rd)  & \red{$\times$}                                                & \red{$\times$}                                                & Unity                             \\
                             & ObMan~\cite{hasson:cvpr19}                            & 2019 & RGB-D    & 256$\times$256   & 154K   & 20 / 3K      & 1 (3rd)  & \red{$\times$}                                                & {\green{\checkmark}}                                          & Blender/GraspIt~\cite{miller:ram05} \\
                             & MVHM~\cite{chen:wacv21}                               & 2021 & RGB-D    & 256$\times$256   & 320K   & - / -       & 8 (3rd)  & \red{$\times$}                                                & \red{$\times$}                                          & Blender                                 \\\rule[0mm]{0mm}{3mm}
\tbf{Hand marker}           & BigHand2.2M~\cite{yuan:cvpr17}                        & 2017 & Depth    & 640$\times$480   & 2,200K   & 10 / -       & 1 (1st + 3rd)  & {\green{\checkmark}}                                          & \red{$\times$}                                          & NDI trakSTAR                      \\
                             & FPHA~\cite{hernando:cvpr18}                           & 2018 & RGB-D    & 640$\times$480 & 105K   & 6 / 4       & 1 (1st)  & {\green{\checkmark}}                                          & {\green{\checkmark}}                                          & NDI trakSTAR                      \\
                             & GRAB~\cite{taheri:eccv20}                             & 2020 & MoCap    & -                & 1,624K  & 10 / 51      & -        & {\green{\checkmark}}                                          & {\green{\checkmark}}                                          & VICON Vantage 16                  \\ \rule[0mm]{0mm}{3mm}
\tbf{Computational}         &                                                       &      &          &                  &        &          &          &                                                               &                                                               &                                   \\
Model fitting               & ICVL~\cite{tang:cvpr14}                  & 2014 & Depth      & 320$\times$240   & 180K    & 10 / -      & 1 (3rd)  & \red{$\times$}                                                & \red{$\times$}                                                & Creative Senz3D                        \\
                             & NYU~\cite{tompson:tog14}                & 2014 & RGB-D       & 640$\times$480   & 81K    & 2 / -      & 1 (3rd)  & \red{$\times$}                                                & \red{$\times$}                                                & PrimeSense \\
                             & FreiHAND~\cite{zimmermann:iccv19}                     & 2019 & RGB      & 224$\times$224   & 37K    & 32 / 27      & 8 (3rd)  & \red{$\times$}                                                & \red{$\times$}                                                & Basler ace                        \\
                             & YouTube3DHands~\cite{kulon:cvpr20}                    & 2020 & RGB      & 640$\times$480   & 47K    & (109) /  -      & 1 (3rd)  & {\green{\checkmark}}                                          &  \red{$\times$}                                               & -                                 \\
                             & HO-3D~\cite{hampali:cvpr20}                           & 2020 & RGB-D    & 640$\times$480   & 103K   & 10 / 10      & 1-5 (3rd)& {\green{\checkmark}}                                          & {\green{\checkmark}}                                          &                                   \\
                             & DexYCB~\cite{chao:cvpr21}                             & 2021 & RGB-D    & 640$\times$480   & 582K   & 10 / 20      & 8 (3rd)  & {\green{\checkmark}}                                          & {\green{\checkmark}}                                          & RealSense D415                    \\\rule[0mm]{0mm}{3mm}
                             & H2O~\cite{kwon:iccv21}                                & 2021 & RGB-D    & 1280$\times$720  & 571K   & 4 / 10      & 5 (1st \& 3rd)  & {\green{\checkmark}}                                   & {\green{\checkmark}}                                          & Azure Kinect                      \\
Triangulation                & Panoptic Studio~\cite{simon:cvpr17}                   & 2017 & RGB      & 1920$\times$1080 & 15K    & - / -       & 31 (3rd)  & {\green{\checkmark}}                                         & \red{$\times$}                                                & HD camera                         \\
                             & InterHand2.6M~\cite{moon:eccv20}                      & 2020 & RGB      & 512$\times$334   & 2,590K  & 27 / -       & 80-140 (3rd) & {\green{\checkmark}}                                      & \red{$\times$}                                                             & HD camera                                  \\
                             & AssemblyHands~\cite{ohkawa:cvpr23}                    & 2023 & RGB/Mono      & 636$\times$480   & 3,030K  & 34 / -       & 12 (1st \& 3rd) & {\green{\checkmark}}                                      & \red{$\times$}                                                             & HD camera/Meta Quest                                  \\
\end{tabular}
}}
\label{tbl:annot}
\end{table*}
}

\def\tblpc{
\begin{table}[tp]
\caption{Pros and cons of each annotation approach.}
\footnotesize{
\scalebox{0.8}[1]{
\begin{tabular}{l|ll}
Annotation    & Pros                                                                                     & Cons                                                                                                        \\ \hline \rule[0mm]{0mm}{3mm}
\tbf{Manual}  & Reasonable accuracy                                                                            & \begin{tabular}[t]{@{}l@{}}Labor intensive\\ Hard to address occlusion\end{tabular}                           \\\rule[0mm]{0mm}{3mm} 
\tbf{Synthetic} & \begin{tabular}[t]{@{}l@{}}Large scale\\ High diveristy\\ Low cost\end{tabular}             & \begin{tabular}[t]{@{}l@{}}Sim-to-real gap\\ Hard to simulate motion\end{tabular}                    \\\rule[0mm]{0mm}{3mm} 
\tbf{Hand marker}  & \begin{tabular}[t]{@{}l@{}}Robust to occlusion\\ Low annotation cost\end{tabular} & \begin{tabular}[t]{@{}l@{}}Requires special sensors\\ Changes visual modality\\ Prevents natural motion\end{tabular}                  \\\rule[0mm]{0mm}{3mm} 
\tbf{Computational} & \begin{tabular}[t]{@{}l@{}}Natural motion\\ Low annotation cost\end{tabular}  & \begin{tabular}[t]{@{}l@{}}Lacks diversity\\ Hard to evaluate quality\\ Needs multi-camera setups\end{tabular}
\end{tabular}
}}
\label{tbl:procon}
\end{table}
}
\begin{document}

\title[aaa]{Efficient Annotation and Learning for 3D Hand Pose Estimation: A Survey}

\author*[1]{\fnm{Takehiko} \sur{Ohkawa}}\email{ohkawa-t@iis.u-tokyo.ac.jp}

\author[1]{\fnm{Ryosuke} \sur{Furuta}}\email{\{furuta, ysato\}@iis.u-tokyo.ac.jp}

\author[1]{\fnm{Yoichi} \sur{Sato}}

\affil[1]{\orgdiv{Institute of Industrial Science}, \orgname{The University of Tokyo}, \orgaddress{\street{4-6-1 Komaba, Meguro-ku}, \city{Tokyo}, \postcode{153-8505}, \country{Japan}}}

\abstract{
In this survey, we present a systematic review of 3D hand pose estimation from the perspective of efficient annotation and learning.
3D hand pose estimation has been an important research area owing to its potential to enable various applications, such as video understanding, AR/VR, and robotics.
However, the performance of models is tied to the quality and quantity of annotated 3D hand poses.
Under the status quo, acquiring such annotated 3D hand poses is challenging, \eg, due to the difficulty of 3D annotation and the presence of occlusion.
To reveal this problem, we review the pros and cons of existing annotation methods classified as manual, synthetic-model-based, hand-sensor-based, and computational approaches.
Additionally, we examine methods for learning 3D hand poses when annotated data are scarce, including self-supervised pretraining, semi-supervised learning, and domain adaptation. 
Based on the study of efficient annotation and learning, we further discuss limitations and possible future directions in this field.
}


\maketitle
\section{Introduction}
The acquisition of 3D hand pose annotations~\footnote{We denote 3D pose as the 3D keypoint coordinates of hand joints, $\mathrm{P}^{\text{3D}} \in \mathbb{R}^{J\times3}$ where $J$ is the number of joints.} has presented a significant challenge in the study of 3D hand pose estimation.
This makes it difficult to construct large training datasets and develop models for various target applications, such as hand-object interaction analysis~\cite{boukhayma:cvpr19,hampali:cvpr20}, pose-based action recognition~\cite{iqbal:fg17,tekin:cvpr19,sener:cvpr22}, augmented and virtual reality~\cite{liang:acmmm15,wu:vcir20,han:tog22}, and robot learning from human demonstration~\cite{ciocarlie:ijrr09,mandikal:corl21,handa:icra20,qin:eccv22}.
In these application scenarios, we must consider methods for annotating hand data, and select an appropriate learning method according to the amount and quality of the annotations.
However, there is currently no established methodology that can give annotations efficiently and learn even from imperfect annotations.
This motivates us to review methods for building training datasets and developing models in the presence of these challenges in the annotation process.

During the annotations, we encounter several obstacles including the difficulty of 3D measurement, occlusion, and dataset bias.
As for the first obstacle, annotating 3D points from a single RGB image is an ill-posed problem.
While annotation methods using hand markers, depth sensors, or multi-view cameras can provide 3D positional labels, these setups require a controlled environment, which limits available scenarios.
As for the second obstacle, occlusion hinders annotators from accurately localizing the positions of hand joints.
As for the third obstacle, annotated data are biased to a specific condition constrained by the annotation method.
For instance, annotation methods based on hand markers or multi-view setups are usually installed in laboratory settings, resulting in a bias toward a limited variety of backgrounds and interacting objects.

\begin{figure}[t]
    \centering
    \includegraphics[width=1.0\linewidth]{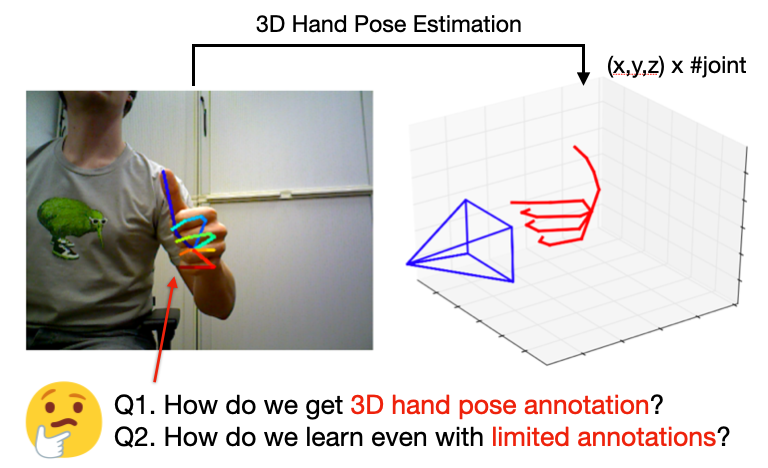}
    \caption{Our survey on 3D hand pose estimation is organized from two aspects: (i) obtaining 3D hand pose annotation and (ii) learning even with a limited amount of annotated data. These two issues will be considered in the scenarios of practical applications where we work on dataset construction and model development with limited resources.
    The figure is adapted from~\cite{zimmermann:iccv17}.
    }
    \label{fig:teaser}
\end{figure}

Given such challenges in annotation, we conduct a systematic review of the literature on 3D hand pose estimation from two distinct perspectives: \textit{efficient annotation} and \textit{efficient learning} (see Fig.~\ref{fig:teaser}).
The former view highlights how existing methods assign reasonable annotations in a cost-effective way, covering a range of topics: the availability and quality of annotations and the limitations when deploying the annotation methods.
The latter view focuses on how models can be developed in scenarios where annotation setups cannot be implemented or available annotations are insufficient.

In contrast to existing surveys on network architecture and modeling~\cite{doosti:arxiv19,le:astes20,liu:icvr21,chatzis:as20,lepetit:arxiv20}, our survey delves into another fundamental direction that arises from the annotation issues, namely, dataset construction with cost-effective annotation and model development with limited resources.
In particular, our survey includes benchmarks, datasets, image capture setups, automatic annotation, learning with limited labels, and transfer learning.
Finally, we discuss potential future directions of this field beyond the current state of the art.

For the study of annotation, we categorize existing methods into manual~\cite{sridhar:eccv16,mueller:iccv17,chao:cvpr21}, synthetic-model-based~\cite{mueller:iccv17,zimmermann:iccv17,mueller:cvpr18,hasson:cvpr19,chen:wacv21}, hand-marker-based~\cite{yuan:cvpr17,hernando:cvpr18,taheri:eccv20}, and computational approaches~\cite{simon:cvpr17,moon:eccv20,zimmermann:iccv19,kulon:cvpr20,hampali:cvpr20,kwon:iccv21}.
While manual annotation requires querying human annotators, hand markers automate the annotation process by tracking sensors attached to a hand.
Synthetic methods utilize computer graphics engines to render plausible hand images with precise keypoint coordinates. 
Computational methods assign labels by fitting a hand template model to the observed data or using multi-view geometry.
We find these annotation methods have their own constraints, such as the necessity of human effort, the sim-to-real gap, the changes in hand appearance, and the limited portability of the camera setups.
Thus, these annotation methods may not always be adopted for every application.

Due to the problems and constraints of each annotation method, we need to consider how to develop models even when we do not have enough annotations.
Therefore, learning with a small amount of labels is another important topic.
For learning from limited annotated data, leveraging a large pool of unlabeled hand images as well as labeled images is a primary interest, \eg, in self-supervised pretraining, semi-supervised learning, and domain adaptation.
Self-supervised pretraining encourages the hand pose estimator to learn from unlabeled hand images, so it enables building a strong feature extractor before performing supervised learning.
While semi-supervised learning trains the estimator with labeled and unlabeled hand images collected from the same environment, domain adaptation further solves the so-called problem of domain gap between the two image sets, \eg, the difference between synthetic data and real data.

\begin{figure*}[t]
    \centering
    \includegraphics[width=0.9\linewidth]{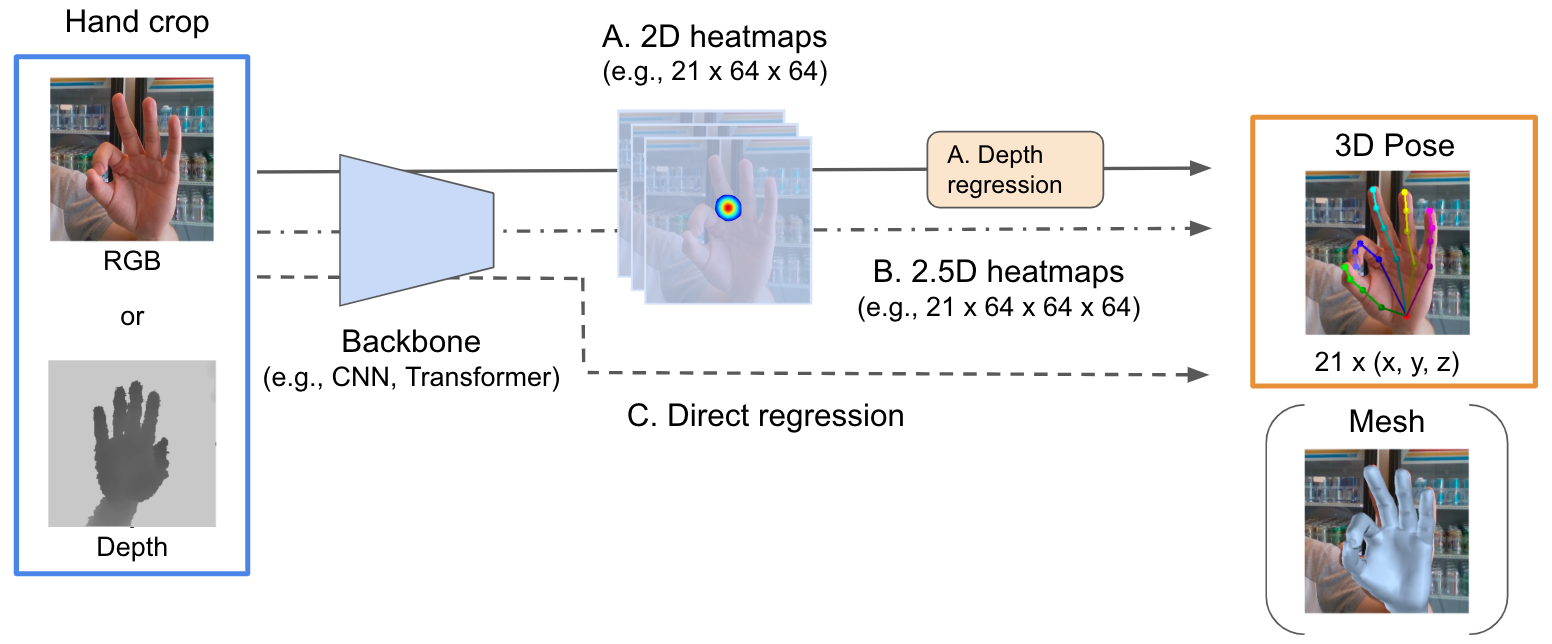}
    \caption{Formulation and modeling of single-view 3D hand pose estimation. For input, we use either RGB or depth images cropped to the hand region. The model learns to produce a 3D hand pose defined by 3D coordinates. Some works additionally estimate hand shape using a 3D hand template model. For modeling, there are three major designs; (A) 2D heatmap regression and depth regression, (B) extended three-dimensional heatmap regression called 2.5D heatmaps, and  (C) direct regression of 3D coordinates.
    }
    \label{fig:modeling}
\end{figure*}

The rest of this survey is organized as follows.
In Section~\ref{sec:overview}, we introduce the formulation and modeling of 3D hand pose estimation.
In Section~\ref{sec:challenges}, we present open challenges in the construction of hand pose datasets involving depth measurement, occlusion, and dataset bias.
In Section~\ref{sec:annotation}, we cover existing methods of 3D hand pose annotation, namely manual, synthetic-model-based, hand-marker-based, and computational approaches.
In Section~\ref{sec:limited_labels}, we provide learning methods from a limited amount of annotated data, namely self-supervised pretraining, semi-supervised learning, and domain adaptation.
In Section~\ref{sec:future}, we finally show promising future directions of 3D hand pose estimation.

\section{Overview of 3D Hand Pose Estimation}\label{sec:overview}
\customparagraph{Task setting.} 
As shown in Fig.~\ref{fig:modeling}, 3D hand pose estimation is typically formulated as the estimation from a monocular RGB/depth image~\cite{erol:cviu07,supancic:ijcv18,yuan:cvpr18}.
The output is parameterized by the hand joint positions with 14, 16, or 21 keypoints, which are introduced in~\cite{tompson:tog14},~\cite{tang:cvpr14}, and~\cite{qian:cvpr14}, respectively.
The dense representation of 21 hand joints~\footnote{Five end keypoints are fingertips, not strictly called joints.} has been popularly used as it contains more precise information about hand structure.
For a single RGB image in which depth and scale are ambiguous, the 3D coordinates of the hand joint relative to the hand root are estimated from a scale-normalized hand image~\cite{zimmermann:iccv17,cai:eccv18,ge:cvpr19}.
Recent works additionally estimate hand shape by regressing 3D hand pose and shape parameters together~\cite{zhou:ijcai16,muller:tog19,ge:cvpr19,boukhayma:cvpr19}.
In evaluation, produced prediction is compared with ground truth, \eg, in the space of world or image coordinates. 
These two metrics are often used: mean per joint position error (MPJPE) in millimeters, and area under curve of percentage of correct keypoints (PCK-AUC).

\customparagraph{Modeling.}
Classic methods estimate a hand pose by finding the closest sample from a large set of hand poses, \eg, synthetic hand pose sets.
Some works formulate the task as nearest neighbor search~\cite{romero:icra10,rogez:iccv15} while others solve pose classification given predefined hand pose classes and a SVM classifier~\cite{rogez:eccvw14,rogez:cvpr15,sridhar:iccv13}.

Recent studies have adopted an end-to-end training manner where models learn the correspondence between the input image and its label of the 3D hand pose.
Standard single-view methods from an RGB image~\cite{zimmermann:iccv17,cai:eccv18,ge:cvpr19} consist of (A) the estimation of 2D hand poses by heatmap regression and depth regression for each 2D keypoint (see Fig.~\ref{fig:modeling}).
The 2D keypoints are learned by optimizing heatmaps centered on each 2D hand joint position.
An additional regression network predicts the depth distance of detected 2D hand keypoints.
Other works use (B) extended 2.5D heatmap regression with a depth-wise heatmap in addition to the 2D heatmaps~\cite{iqbal:eccv18,moon:eccv20}, so it does not require a depth regression branch.
Depth-based hand pose estimation also utilizes such heatmap regression~\cite{huang:aaai20,xiong:iccv19,ren:bmvc19}.
Instead of the heatmap training, other methods learn to (C) directly regress keypoint coordinates~\cite{spurr:cvpr18,santavas:sensors21}.

For the architecture of the backbone network, CNNs (\eg, ResNet~\cite{he:cvpr16}) are a basic choice while recent Transformer-based methods have been proposed~\cite{hampali:cvpr22,huang:eccv20}.
To generate feasible hand poses, regularization is a key trick in correcting predicted 3D hand poses.
Based on the anatomical study of hands, bio-mechanical constraints are imposed to limit predicted bone lengths and joint angles~\cite{spurr:eccv20,chen:cvpr21,liu:cvpr21}.

\section{Challenges in Dataset Construction}\label{sec:challenges}
Task formulation and algorithms for estimating 3D hand poses are outlined in Section~\ref{sec:overview}.
During training, it is necessary to build a large amount of training data with diverse hand poses, viewpoints, and backgrounds.
However, obtaining such massive hand data with accurate annotations has been challenging for the following reasons.

\begin{figure}[tp]
    \centering
    \includegraphics[width=1.0\linewidth]{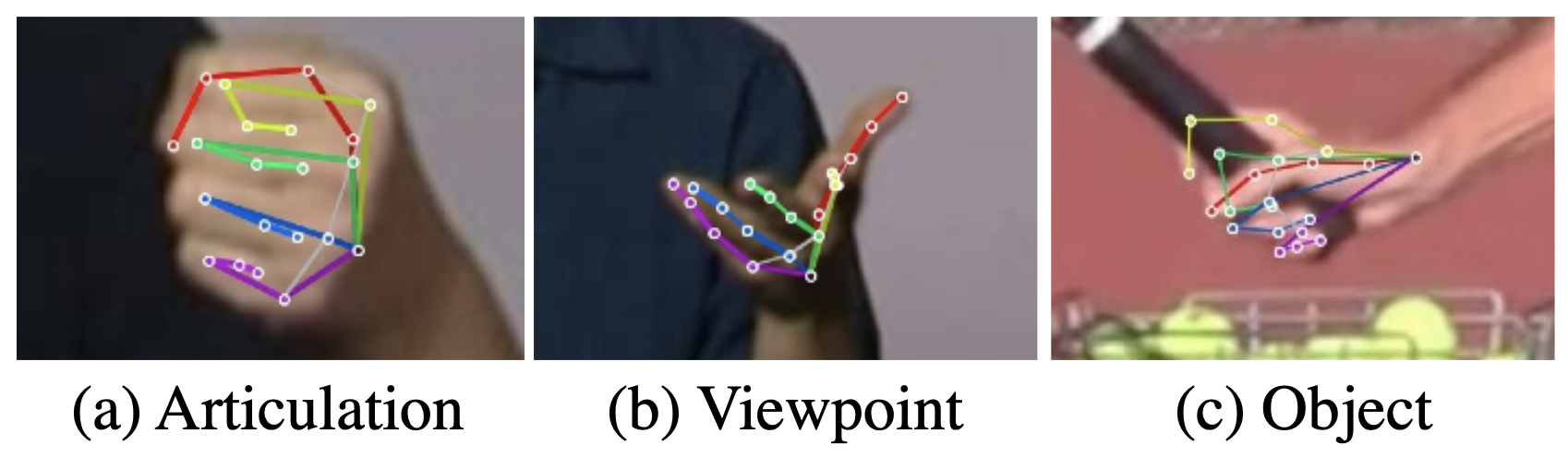}
    \caption{Difficulty of hand pose annotation in a single RGB image~\cite{simon:cvpr17}. Occlusion of hand joints is caused by (a) articulation, (b) viewpoint bias, and (c) grasping objects.}
    \label{fig:hand_annot}
\end{figure}

\customparagraph{Difficulty of 3D annotation.}
Annotating the 3D position of hand joints from a single RGB image is inherently impossible without any prior information or additional sensors due to an ill-posed condition. 
To assign accurate hand pose labels, hand-marker-based annotation using magnetic sensors~\cite{wetzler:bmvc15,yuan:cvpr17,hernando:cvpr18}, motion capture systems~\cite{miyata:iros04,schroder:mig15,taheri:eccv20}, or hand gloves~\cite{wang:tog09,bianchi:ijrr13,glauser:tog19} has been studied.
These sensors can provide 6-DoF information (\ie, location and orientation) of attached markers and enable us to calculate the coordinates of full hand joints from the tracked markers.
However, their setups are expensive and need good calibration, which constrains available scenarios.

On the contrary, depth sensors (\eg, RealSense) or multi-view camera studios~\cite{simon:cvpr17,moon:eccv20,chao:cvpr21,zimmermann:iccv19,hampali:cvpr20} make it possible to obtain depth information near hand regions.
Given 2D keypoints for an image, these setups enable annotation of 3D hand poses by measuring the depth distance at each 2D keypoint.
However, these annotation methods do not always produce satisfactory 3D annotations, \eg, due to an occlusion problem (detailed in the next section).
In addition, depth images are significantly affected by the sensor noise, 
such as unknown depth values in some regions and ghost shadows around object boundaries~\cite{xu:iccv13}.
Due to the limited depth distance that depth cameras can capture, the depth measurement becomes inaccurate when the hands are far from the sensor.

\customparagraph{Occlusion.}
Hand images often contain complex occlusions that distract human annotators from localizing hand keypoints.
Examples of possible occlusions are shown in Fig.~\ref{fig:hand_annot}.
In figure (a), articulation causes a self-occlusion that makes some hand joints (\eg, fingertips) invisible due to the overlap with the other parts of the hand. 
In figure (b), such self-occlusion depends on a specific camera viewpoint.
In figure (c), hand-held objects induce occlusion that hides the hand joints by the object during the interaction. 

To address this issue, hand-marker-based tracking~\cite{wetzler:bmvc15,yuan:cvpr17,hernando:cvpr18,taheri:eccv20} 
and multi-view camera studios~\cite{simon:cvpr17,moon:eccv20,chao:cvpr21,zimmermann:iccv19,hampali:cvpr20} have been studied. 
The hand markers offer 6-DoF information during these occlusions, so the hand-maker-based annotation is robust to the occlusion.
For multi-camera settings, the effect of occlusion can be reduced when many cameras are densely arranged.

\begin{figure}[tp]
    \centering
    \includegraphics[width=1.0\linewidth]{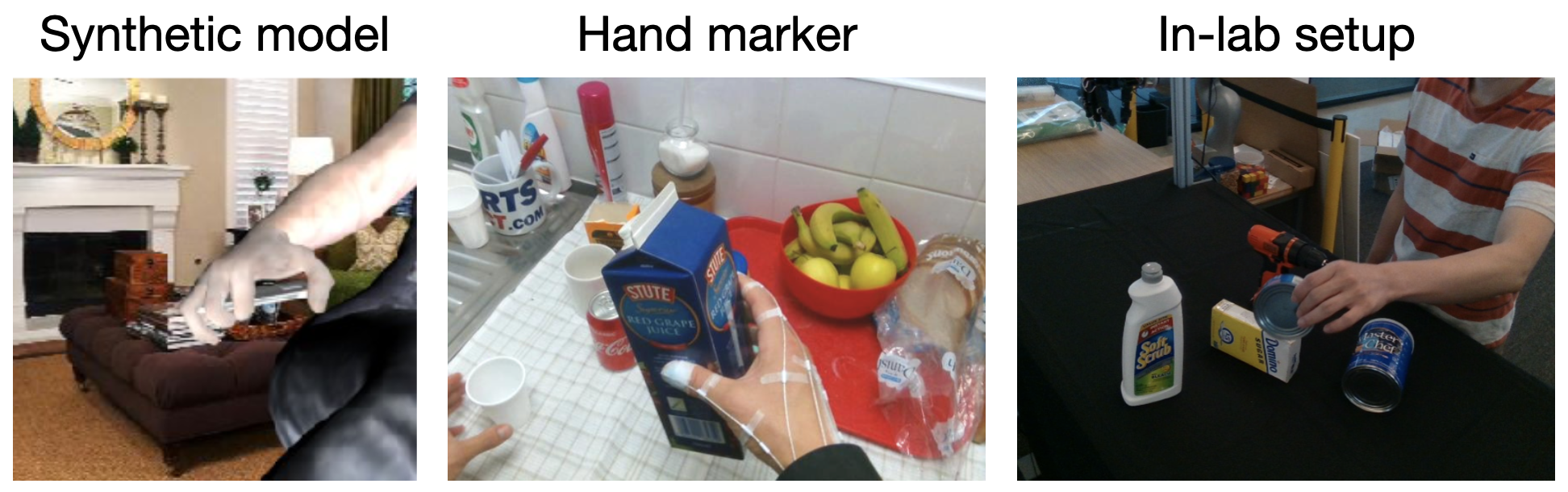}
    \caption{Example of major data collection setups.
    The synthetic image on the left (ObMan~\cite{hasson:cvpr19}) can be generated inexpensively, but they exhibit unrealistic hand texture.
    The hand markers on the middle (FPHA~\cite{hernando:cvpr18}) enable automatic tracking of hand joints, although the markers distort the appearance of hands. 
    The in-lab setup on the right (DexYCB~\cite{chao:cvpr21}) uses a black background to make it easier to recognize hands and objects, but it limits data variation in environments.
    }
    \label{fig:dataset_bias}
\end{figure}

\tblannot

\customparagraph{Dataset bias.}
While hands are a common entity in various image capture settings, the category of objects, including hand-held objects (\ie, foregrounds) and backgrounds, is potentially diverse.
In order to improve the generalization ability of hand pose estimators, hand images must be annotated under various imaging conditions (\eg, lighting, viewpoints, hand poses, and backgrounds).
However, it is challenging to create such large and diverse datasets nowadays due to the aforementioned problems.
Rather, existing hand pose datasets exhibit a bias to a particular imaging condition constrained by the annotation method.

As shown in Fig.~\ref{fig:dataset_bias}, generating data using synthetic models~\cite{mueller:iccv17,zimmermann:iccv17,mueller:cvpr18,hasson:cvpr19,chen:wacv21} is cost-effective, but it creates unrealistic hand texture~\cite{ohkawa:access21}.
Although the hand-marker-based annotation~\cite{wetzler:bmvc15,yuan:cvpr17,hernando:cvpr18,taheri:eccv20} can automatically track the hand joints from the information of hand sensors, the sensors distort the hand appearance and hinder the natural hand movement.
In-lab data acquired by multi-camera setups~\cite{simon:cvpr17,moon:eccv20,chao:cvpr21,zimmermann:iccv19,hampali:cvpr20} make the annotation easier because they can reduce the occlusion effect.
However, the variations in environments (\eg, backgrounds and interacting objects) are limited because the setups are not easily portable.

\section{Annotation Methods}\label{sec:annotation}
Given the above challenges concerning the construction of hand pose datasets, we review existing 3D hand pose datasets in terms of annotation design.
As shown in Table~\ref{tbl:annot}, we categorize the annotation methods as manual, synthetic-model-based, hand-marker-based, and computational approaches.
We then study the pros and cons of each annotation method in Table~\ref{tbl:procon}.

\subsection{Manual annotation}
MSRA~\cite{qian:cvpr14}, Dexter+Object~\cite{sridhar:eccv16}, and EgoDexter~\cite{mueller:iccv17} manually annotate 2D hand keypoints on the depth images and determine the depth distance from the depth value of the images on the 2D point.
This method enables assigning reasonable annotations of 3D coordinates (\ie, 2D position and depth) when hand joints are fully visible.

However, it is not extensively available according to the number of frames due to the high annotation cost.
In addition, since it is not robust for occluded keypoints, this approach only allows fingertip annotation, instead of full hand joints.
For these limitations, these datasets provide a small amount of data ($\approx 3\text{K}$ images) used for evaluation only.
Additionally, these single-view datasets can produce view-dependent annotation errors because a single-depth camera captures the distance to the hand skin surface, not the true joint position.
To reduce such unavoidable errors, subsequent annotation methods based on multi-camera setups provide further accurate annotations (see Section~\ref{sec:comp-annot}).

\tblpc

\subsection{Synthetic-model-based annotation}
To acquire large-scale hand images and labels, synthetic methods based on synthetic hand and full-body models~\cite{rogez:eccvw14,libhand,loper:tog15,romero:tog17} have been proposed.
SynthHands~\cite{mueller:iccv17} and RHD~\cite{zimmermann:iccv17} render synthetic hand images with randomized real backgrounds from either a first- or third-person view.
MVHM~\cite{chen:wacv21} generates multi-view synthetic hand data rendered from eight viewpoints.
These datasets have succeeded in providing accurate hand keypoint labels on a large scale.
Although they can generate various background patterns inexpensively, the lighting and texture of hands are not well simulated, and the simulation of hand-object interaction is not considered in the data generation process.

To handle these issues, GANerated~\cite{mueller:cvpr18} utilizes GAN-based image translation to stylize synthetic hands more realistically.
Furthermore, ObMan~\cite{hasson:cvpr19} simulates the hand-object interaction in data generation using a hand grasp simulator (Graspit~\cite{miller:ram05}) with known 3D object models (ShapeNet~\cite{chang:arxiv15}).
Ohkawa~\etal proposed foreground-aware image stylization to convert the simulation texture in the ObMan data to a more realistic one while separating the hand regions and backgrounds~\cite{ohkawa:access21}.
Corona~\etal attempted to synthesize more natural hand grasps with affordance classification and the refinement of fingertip locations~\cite{corona:cvpr20}.
However, the ObMan data only provide static hand images with hand-held objects, not hand motion.
The hand motion simulation while approaching the object remains an open problem.

\begin{figure}[tp]
    \centering
    \includegraphics[width=1\linewidth]{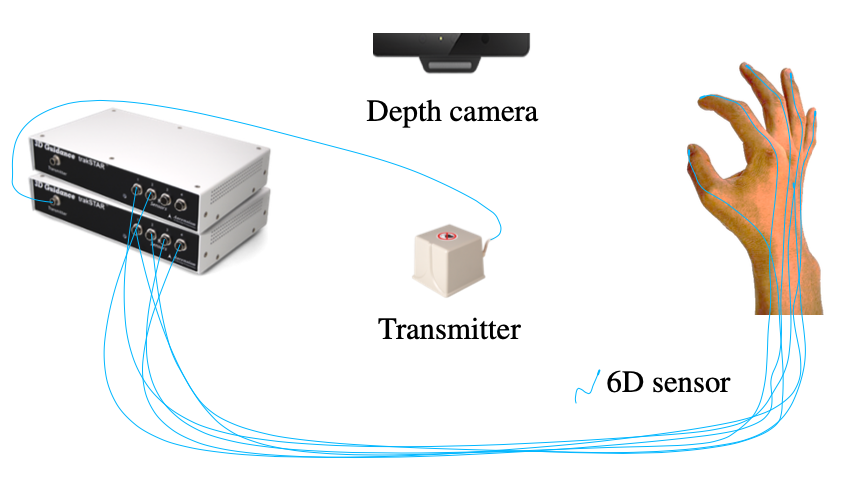}
    \caption{Illustration of a hand marker setup~\cite{yuan:cvpr17}.}
    \label{fig:sensor}
\end{figure}

\begin{figure}[tp]
    \centering
    \includegraphics[width=0.8\linewidth]{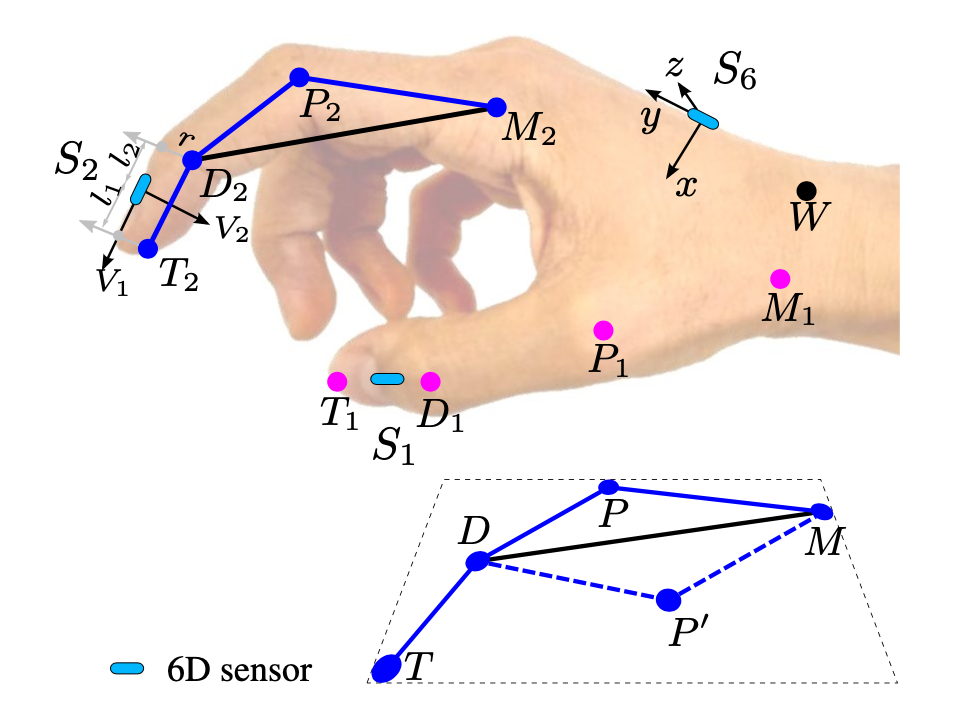}
    \caption{Calculation of joint positions from tracked markers~\cite{yuan:cvpr17}. $S_i$ denotes the position of the markers, and $W$, $M_i$, $P_i$, $D_i$, and $T_i$ are the positions of hand joints listed from the wrist to the fingertips.
    }
    \label{fig:marker}
\end{figure}

\subsection{Hand-marker-based annotation}
As shown in Fig.~\ref{fig:sensor}, hand-marker-based annotation automatically 
tracks attached hand markers and then calculates the coordinates of hand joints.
Initially, Wetzler~\etal attached magnetic sensors to fingertips that provide 6-DoF information of the markers~\cite{wetzler:bmvc15}.
While this scheme can annotate fingertips only, recent datasets, BigHand2.2M~\cite{yuan:cvpr17} and FPHA~\cite{hernando:cvpr18}, use these sensors to offer the annotation of the full 21 hand joints.
Fig.~\ref{fig:marker} shows how to compute the joint positions given six magnetic sensors.
It uses inverse kinematics to infer all 21 hand joints, which fits a hand skeleton with the constraints of the maker positions and user-specific bone length manually measured beforehand.

However, these sensors obstruct natural hand movement and distort the appearance of the hand.
Due to the changes in hand appearance, these datasets have been proposed for the benchmark of depth-based estimation, not the RGB-based task.
On the contrary, GRAB~\cite{taheri:eccv20} is built with a motion capture system for human hands and body, but it does not possess visual modality, \eg, RGB images.

\begin{figure}[tp]
    \centering
    \includegraphics[width=0.9\linewidth]{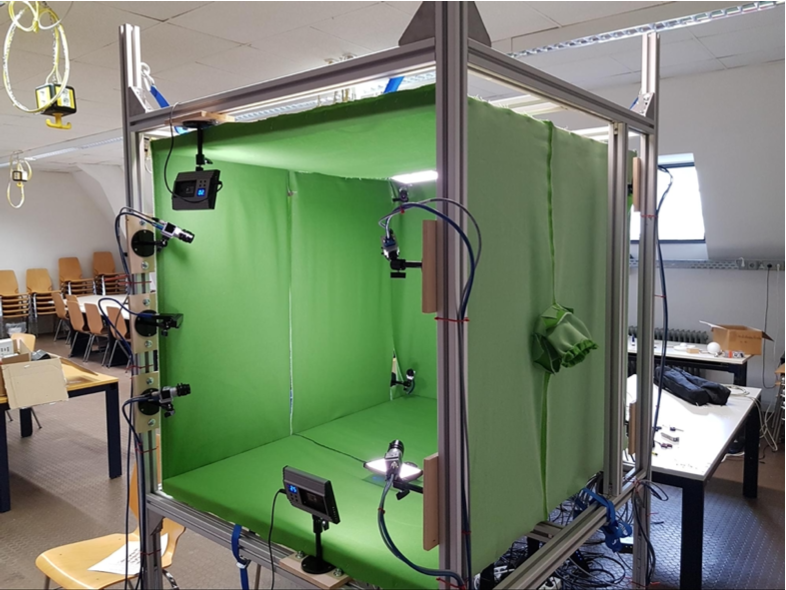}
    \caption{Illustration of a multi-camera setup~\cite{zimmermann:iccv19}.}
    \label{fig:multiview}
\end{figure}

\subsection{Computational annotation}\label{sec:comp-annot}
Computational annotation is categorized into two major approaches: \tit{hand model fitting} and \tit{triangulation}.
Unlike hand-marker-based annotation, these methods can capture natural hand motion without attaching hand markers.

\customparagraph{Model fitting (depth).} Early works of computational annotation utilize \textit{model fitting} on depth images~\cite{supancic:ijcv18,yuan:cvpr18}.
Since a depth image provides 3D structural information, their works fit a 3D hand model, from which joint positions can be obtained, to the depth image.
ICVL~\cite{tang:cvpr14} fits a convex rigid body model by solving a linear complementary problem with physical constraints~\cite{melax:gi13}.
NYU~\cite{tompson:tog14} uses a hand model defined by spheres and cylinders and formulates the model fitting as a kind of particle swarm optimization~\cite{oikonomidis:bmvc11,oikonomidis:cvpr12}.
The use of other cues for the model fitting is also studied~\cite{lu:cvpr03,ballan:eccv12}, such as edges, optical flow, shading, and collisions.
Sharp~\etal paint hands to obtain hand part labels by color segmentation on RGB images and the proxy cue of hand parts further helps the depth-based model fitting~\cite{sharp:chi15}.

Using these depth datasets, several more accurate labeling methods have been proposed.
Rogez~\etal gave manual annotation to a few joints and searched the closest 3D pose from a pool of synthetic hand pose data~\cite{rogez:eccvw14}.
Oberweger~\etal considered model fitting with temporal coherence~\cite{oberweger:cvpr16}.
This method selects reference frames from a depth video and asks annotators for manual labeling.
Model fitting is done separately for annotated reference frames and unlabeled non-reference frames.
Finally, all sequential poses are optimized to satisfy temporal smoothness.

\begin{figure}[tp]
    \centering
    \includegraphics[width=0.9\linewidth]{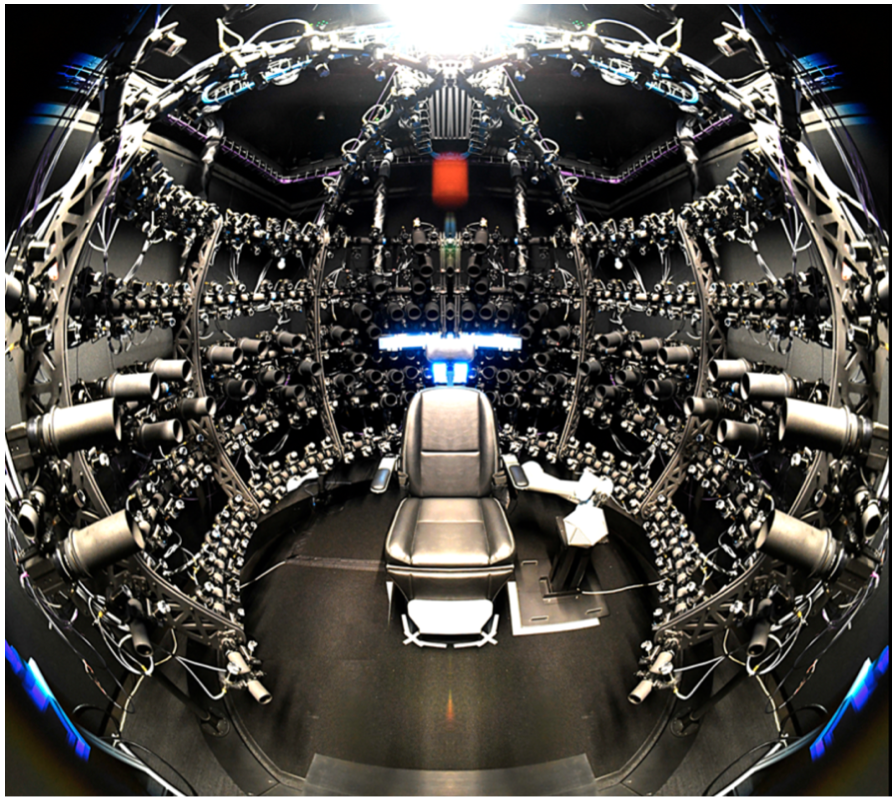}
    \caption{Illustration of a many-camera setup~\cite{wuu:arxiv22}. This setup has about 100 synchronized cameras and is used to create the InterHands2.6M dataset~\cite{moon:eccv20}.
    }
    \label{fig:mugsy}
\end{figure}

\customparagraph{Triangulation (RGB).} For the annotation of RGB images, a multi-camera studio is often used to compute 3D points by multi-view geometry, \ie, \textit{triangulation} (see Fig.~\ref{fig:multiview}).
Panoptic Studio~\cite{simon:cvpr17} and InterHand2.6M~\cite{moon:eccv20} triangulate a 3D hand pose from multiple 2D hand keypoints provided by an open source library, OpenPose~\cite{openpose}, or human annotators.
The generated 3D hand pose is reprojected onto the image planes of other cameras to annotate hand images with novel viewpoints.
This multi-view annotation scheme is beneficial when many cameras are installed (see Fig.~\ref{fig:mugsy}).
For instance, the InterHand2.6M manually annotates keypoints from 6 views and reprojects the triangulated points to the other many views (100+). 
This setup can produce over 100 training images for every single annotation.
The InterHand2.6M has million-scale training data.

This point-level triangulation method works quite well when many cameras (30+) are arranged~\cite{simon:cvpr17,moon:eccv20}.
However, the AssemblyHands setup~\cite{ohkawa:cvpr23} has only eight static cameras, and then the predicted 2D keypoints to be triangulated tend to be suboptimal due to hand-object occlusion during the assembly task.
To improve the accuracy of triangulation in such sparse camera settings, Ohkawa~\etal adopt multi-view aggregation of encoded features by the 2D keypoint detector and compute 3D coordinates from constructed 3D volumetric features~\cite{ohkawa:cvpr23,zimmermann:iccv19,bartol:cvpr22,iskakov:iccv19}.
This feature-level triangulation provides better accuracy than the point-level method, achieving an average keypoint error of 4.20~mm, which is 85\% lower than the error of the original annotations in Assembly101~\cite{sener:cvpr22}.

\customparagraph{Model fitting (RGB).} Model fitting is also used in RGB-based pose annotation.
FreiHAND~\cite{zimmermann:iccv19,zimmermann:gcpr21} utilizes a 3D hand template (MANO~\cite{romero:tog17}) fitting to multi-view hand images with sparse 2D keypoint annotation.
The dataset increases the variation of training images by randomly synthesizing the background and using captured real hands as the foreground.
YouTube3DHands~\cite{kulon:cvpr20} uses the MANO model fitting to estimated 2D hand poses in YouTube videos.
HO-3D~\cite{hampali:cvpr20}, DexYCB~\cite{chao:cvpr21}, and H2O~\cite{kwon:iccv21} jointly annotate 3D hand and object poses to facilitate a better understanding of hand-object interaction.
Using estimated or manually annotated 2D keypoints, their datasets fit the MANO model and 3D object models to the hand images with objects.

\begin{figure}[tp]
    \centering
    \includegraphics[width=1\linewidth]{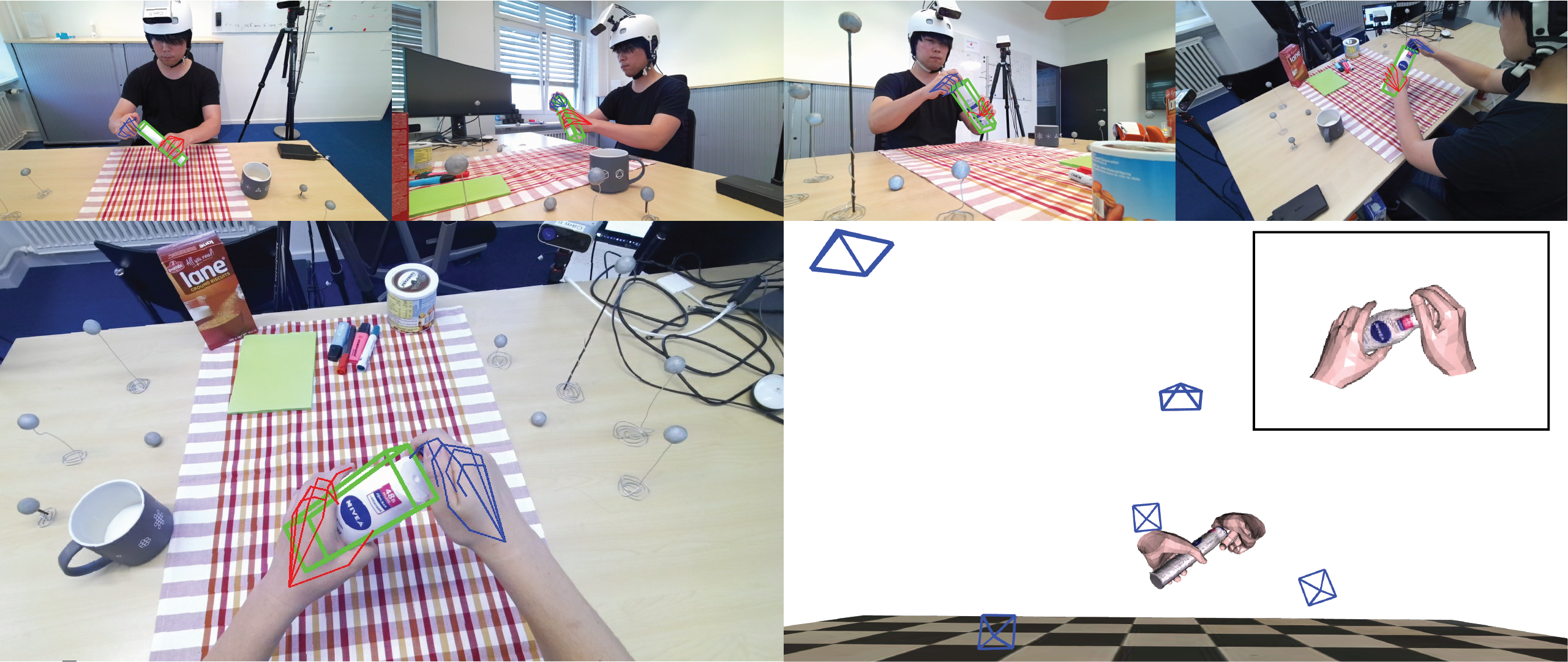}
    \caption{
    Synchronized multi-camera setup with first-person and third-person cameras~\cite{kwon:iccv21}.
    }
    \label{fig:ego-exo}
\end{figure}

While most methods capture hands from static third-person cameras, H2O and AssemblyHands install first-person cameras that are synchronized with static third-person cameras (see Fig.~\ref{fig:ego-exo}).
With camera calibration and head-mounted camera tracking, such camera systems can offer 3D hand pose annotations for first-person images by projecting annotated keypoints from third-person cameras onto first-person image planes.
This reduces the cost of annotating first-person images, which is considered expensive because the image distribution changes drastically over time and the hands are sometimes out of view.

These computational methods can generate labels with little human effort, although the camera system itself is costly.
However, assessing the quality of the labels is still difficult.
In fact, the annotation quality depends on the number of cameras and their arrangement, the accuracy of hand detection and the estimation of 2D hand poses, and the performance of triangulation and fitting algorithms.

\section{Learning with Limited Labels}\label{sec:limited_labels}
As explained in Section~\ref{sec:annotation}, existing annotation methods have certain pros and cons.
Since perfect annotation in terms of amount and quality cannot be assumed, training 3D hand pose estimators with limited annotated data is another important study.
Accordingly, we introduce learning methods using unlabeled data in this section, namely self-supervised pretraining, semi-supervised learning, and domain adaptation.

\begin{figure}[t]
    \centering
    \includegraphics[width=1\linewidth]{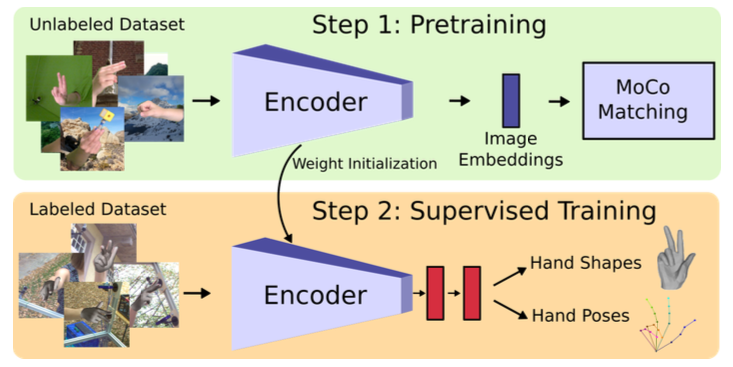}
    \caption{Self-supervised pretraining of 3D hand pose estimation~\cite{zimmermann:gcpr21}. The pretraining phase (step 1) aims to construct an improved encoder network by using many unlabeled data before supervised learning (step 2). The work uses MoCo~\cite{he:cvpr20} as a method of self-supervised learning.
    }
    \label{fig:hand_ssl}
\end{figure}

\subsection{Self-supervised pretraining and learning}
Self-supervised pretraining aims to utilize massive unlabeled hand images and build an improved encoder network before supervised learning with labeled images.
As shown in Fig.~\ref{fig:hand_ssl}, recent works~\cite{spurr:iccv21,zimmermann:gcpr21} first pretrain an encoder network that extracts image features by using contrastive learning (\eg, MoCo~\cite{he:cvpr20} and SimCLR~\cite{chen:icml20}) and then fine-tune the whole network in a supervised manner.
The core idea of contrastive learning is to push a pair of \tit{similar} instances closer together in an embedding space while unrelated instances are pushed apart. 
This approach focuses on how to define the \tit{similarity} of hand images and how to design embedding techniques.
Spurr~\etal proposed to geometrically align two features generated from differently augmented instances~\cite{spurr:iccv21}.
Zimmermann~\etal found that multi-view images representing the same hand pose can be effective pair supervision~\cite{zimmermann:gcpr21}.

Other works utilize the scheme of self-supervised learning that solves an auxiliary task, instead of the target task of hand pose estimation.
Given the prediction on an unlabeled depth image, \cite{wan:cvpr19,oberweger:iccv15}~render a synthetic depth image and penalize the matching between the input image and the one generated from the prediction.
This auxiliary loss by image synthesis is informative even when annotations are scarce.

\begin{figure}[t]
    \centering
    \includegraphics[width=1\linewidth]{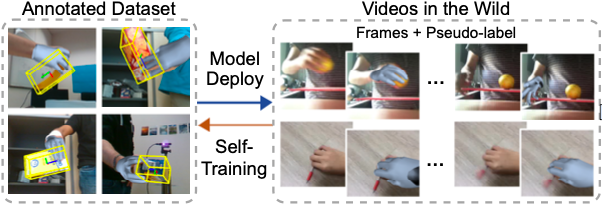}
    \caption{Semi-supervised learning of 3D hand pose estimation~\cite{liu:cvpr21}. The model is trained jointly on annotated data and unlabeled data with pseudo-labels.}
    \label{fig:hand_semi}
\end{figure}

\begin{figure*}[t]
    \begin{tabular}{cc}
      \begin{minipage}[t]{0.48\linewidth}
        \centering
        \includegraphics[width=1\linewidth]{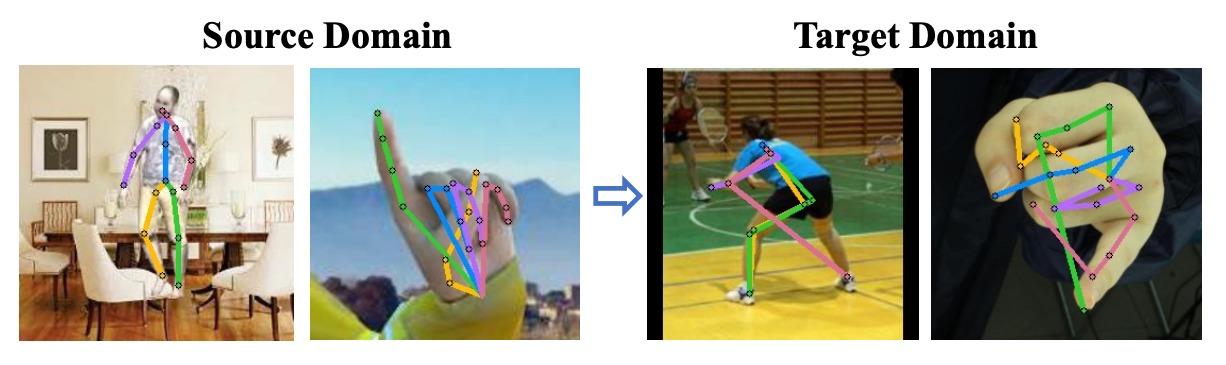}
        \caption{Poor generalization to an unknown domain~\cite{jiang:cvpr21}. The models trained on synthetic images (source) exhibit a limited capacity for inferring poses on real images (target).}
        \label{fig:sim2real}
      \end{minipage} &
      \begin{minipage}[t]{0.48\linewidth}
        \centering
        \includegraphics[width=1\linewidth]{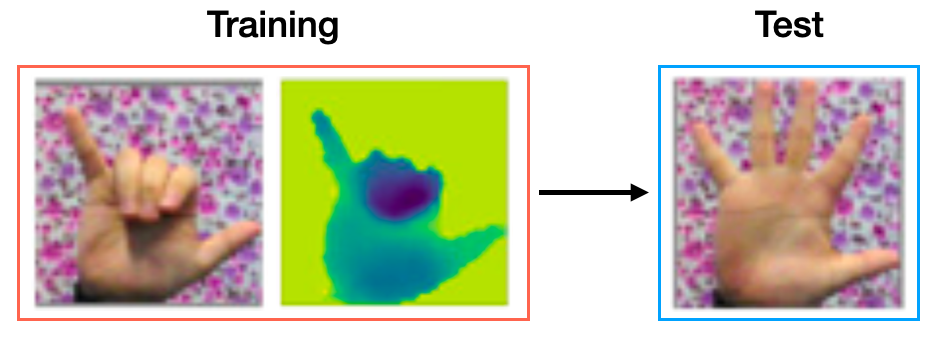}
        \caption{Example of modality transfer. During training, RGB and depth images are accessible and RGB images are given in the test phase. The training aims to utilize the support of depth information to improve RGB-based hand pose estimation. }
        \label{fig:modality}
      \end{minipage}
    \end{tabular}
  \end{figure*}

\subsection{Semi-supervised learning}
As shown in Fig.~\ref{fig:hand_semi}, semi-supervised learning is used to learn from small labeled data and large unlabeled data simultaneously.
Liu~\etal proposed a pseudo-labeling method that learns unlabeled instances with pseudo-ground-truth given from the model's prediction~\cite{liu:cvpr21}.
This pseudo-label training is applied only when its prediction satisfies spatial and temporal constraints.
The spatial constraints check the correspondence of a 2D hand pose
and the 2D pose projected from 3D hand pose prediction.
In addition, they include a constraint based on bio-mechanical feasibility, such as bone lengths and joint angles.
The temporal constraints indicate the smoothness of hand pose and mesh predictions over time.

Yang~\etal proposed the combination of pseudo-labeling and consistency training~\cite{yang:iccv21}.
In pseudo-labeling, the generated pseudo-labels are corrected by fitting the hand template model.
In addition, the work enforces consistency losses between the predictions of differently augmented instances and between the modalities of 2D hand poses and hand masks.

Spurr~\etal applied adversarial training to a sequence of predicted hand poses~\cite{spurr:arxiv21}.
The encoder network is expected to be improved by fooling a discriminator that distinguishes between plausible and invalid hand poses.

\subsection{Domain adaptation}
Domain adaptation aims to improve model performance on target data by learning from labeled source data and target data with limited labels.
This study has addressed two types of underlying domain gaps: \textit{between different datasets} and \textit{between different modalities}.

The former problem \textit{between different datasets} is a common domain adaptation problem where the source and target data are sampled from two datasets with different image statistics, \eg, sim-to-real adaptation~\cite{tang:iccv13,jiang:cvpr21} (see Fig.~\ref{fig:sim2real}).
The model has access to readily available synthetic images with labels and target real images without labels.
The latter problem \textit{between different modalities} is characterized as \textit{modality transfer} where the source and target data represent the same scene, but their modalities are different, \eg, depth \vs RGB (see Fig.~\ref{fig:modality}).
This aims to utilize information-rich source data, \eg, depth images contain 3D structural information, for inferring easily available target data (\eg, RGB images).

To reduce the gap between the two datasets, two major approaches have been proposed: \textit{generative methods} and \textit{adversarial methods}.
In generative methods, Qi~\etal proposed an image translation method to alter the synthetic textures to realistic ones and train a model on generated real-like synthetic data~\cite{qi:arxiv20}.

Adversarial methods enforce matching two domains' features so that the feature extractor can encode features even from the target domain.
However, in addition to the domain gap in an input space (\eg, the difference in backgrounds), the gap in a label space also exists in this task, which is not assumed in typical adversarial methods~\cite{ganin:icml15, tzeng:cvpr17}.
Zhang~\etal developed a feature matching method based on Wasserstein distance and proposed adaptive weighting to enable matching only for features related to hand characteristics, except for label information~\cite{zhang:mm20}.
Jiang~\etal utilized an adversarial regressor and optimized the domain disparity by a minimax game~\cite{jiang:cvpr21}.
Such minimax of disparity is effective in domain adaptation of regression tasks, including hand pose estimation.

As for the \textit{modality transfer} problem, Yuan~\etal and Rad~\etal attempted to use depth images as the auxiliary information during training and test the model on RGB images~\cite{yuan:iccvw19,rad:accv18}.
They observed that learned features from depth images could support RGB-based hand pose estimation.
Park~\etal transferred the knowledge from depth images to infrared (IR) images that have less motion blur~\cite{park:ismar20}. 
Their training is facilitated by matching two features from paired images, \eg, (RGB, depth) and (depth, IR).
Baek~\etal newly defined the domain of hand-only images where a hand-held object is removed.
The work translates hand-object images to hand-only images by using GAN and mesh renderer~\cite{baek:cvpr20}.
Given a hand-object image with an unknown object, this method can generate hand-only images, from which hand pose estimation is more tractable.

\section{Future Directions}\label{sec:future}
\subsection{Flexible camera systems}
We believe that hand image capture will feature more flexible camera systems, such as using first-person cameras.
To reduce the occlusion effect without the need for hand markers, recently published hand datasets have been acquired by multi-camera setups, \eg, DexYCB~\cite{chao:cvpr21}, InterHand2.6M~\cite{moon:eccv20}, and FreiHAND~\cite{zimmermann:iccv19}.
These setups are static and not suitable for capturing dynamic user behavior.
To address this, a \textit{first-person camera} attached to the user's head or body is useful because it mostly captures close-up hands even when the user moves around.
However, as shown in Table~\ref{tbl:annot}, existing first-person benchmarks have a very limited variety due to heavy occlusion, motion blur, and a narrow field-of-view.

One promising direction is a joint camera setup with first-person and third-person cameras, such as H2O~\cite{kwon:iccv21} and AssemblyHands~\cite{ohkawa:cvpr23}.
This results in flexibly capturing the user's hands from the first-person camera while taking the benefits of multiple third-person cameras (\eg, mitigating the occlusion effect).
However, the first-person camera wearer doesn't always have to be alone.
Image capture with multiple first-person camera wearers in a static camera setup will advance the analysis of multi-person cooperation and interaction, \eg, game playing and construction with multiple people.

\subsection{Various types of activities}
We believe that increasing the type of activities is an important direction for generalizing models to various situations with hand-object interaction.
A major limitation of existing hand datasets is the narrow variation of users' performing tasks and grasping objects.
To avoid object occlusion, some works did not capture hand-object interaction~\cite{zimmermann:iccv17,yuan:cvpr17,moon:eccv20}.
Others~\cite{chao:cvpr21,hasson:cvpr19,hampali:cvpr20} used pre-registered 3D object models (\eg, YCB~\cite{berk:ram15}) to simplify in-hand object pose estimation.
User action is also very simple in these benchmarks, such as \textit{pick and place}.

From an affordance perspective~\cite{hassanin:compsurvey21}, diversifying the object category will result in increasing hand pose variation.
Potential future works will capture goal-oriented and procedural activities that naturally occur in our daily life~\cite{grauman:cvpr22,damen:ijcv21,sener:cvpr22}, such as cooking, art and craft, and assembly.

To enable this, we need to develop portable camera systems and robust annotation methods for complex backgrounds and unknown objects.
In addition, occurring hand poses are constrained to the context of the activity.
Thus, pose estimators conditioned by actions, objects, or textual descriptions of the scene will improve estimation in various activities.

\subsection{Towards minimal human effort}
Sections~\ref{sec:annotation} and~\ref{sec:limited_labels} separately explain efficient annotation and learning.
To minimize the effort of human intervention, utilizing findings from both annotation and learning perspectives is one of the promising directions.
Feng~\etal exploited \tit{active learning} that optimizes which unlabeled instance should be annotated and semi-supervised learning that jointly utilizes labeled data and large unlabeled data~\cite{feng:arxiv21}.
However, this method is constrained to triangulation-based 3D pose estimation.
As we mentioned in Section~\ref{sec:comp-annot}, another major computational annotation is \textit{model fitting}; thus, we still need to consider such a collaborative approach in the annotation based on model fitting.

Zimmermann~\etal also proposed a framework of human-in-loop annotation that inspects the annotation quality manually while updating annotation networks on the inspected annotations~\cite{zimmermann:iccv17}.
However, this human check will be a bottleneck in large dataset construction. 
The evaluation of annotation quality on the fly is a necessary technique to scale up the combination of annotation and learning.

\subsection{Generalization and adaptation}
Increasing the generalization ability across different datasets or adapting models to a specific domain is a remaining issue.
The bias of existing training datasets hinders the estimators from inferring test images captured under very different imaging conditions.
In fact, as reported in~\cite{zimmermann:iccv19,han:tog20}, models trained on existing hand pose datasets poorly generalize to other datasets.
For real-world applications (\eg, AR), it is crucial to transfer models from indoor hand datasets to outdoor videos because common multi-camera setups are not available outdoors~\cite{ohkawa:eccv22}.
Thus, aggregating multiple annotated yet biased datasets for generalization and robustly adapting to very different environments are important future tasks.

\section{Summary}
We presented the survey of 3D hand pose estimation from the standpoint of efficient annotation and learning.
We provided a comprehensive overview of this task and modeling, and open challenges during dataset construction.
We investigated annotation methods categorized as manual, synthetic-model-based, hand-marker-based, and computational approaches, and examined their respective strengths and weaknesses.
In addition, we studied learning methods that can be applied even when annotations are scarce, namely self-supervised pretraining, semi-supervised learning, and domain adaptation.
Finally, we discussed potential future advancements in 3D hand pose estimation, including next-generation camera setups, increased object and action variation, jointly optimized annotation and learning techniques, and generalization and adaptation.

\backmatter

\bibliographystyle{ieee_fullname}
\bibliography{main.bbl}


\end{document}